# Strategic Reflectivism In Intelligent Systems

Nick Byrd[1][0000-0001-5475-5941]

[1] Geisinger College of Health Sciences, Danville PA 17822, USA
byrdnick.com/contact

**Abstract.** By late 20th century, the rationality wars had launched debates about the nature and norms of intuitive and reflective thinking. Those debates drew from mid-20th century ideas such as bounded rationality, which challenged more idealized notions of rationality observed since the 19th century. Now that 21st century cognitive scientists are applying the resulting dual process theories to artificial intelligence, it is time to dust off some lessons from this history. So this paper synthesizes old ideas with recent results from experiments on humans and machines. The result is *Strategic Reflectivism*, the position that one key to intelligent systems (human or artificial) is pragmatic switching between intuitive and reflective inference to optimally fulfill competing goals. Strategic Reflectivism builds on American Pragmatism, transcends superficial indicators of reflective thinking such as model size or chains of thought, applies to both individual and collective intelligence systems (including human-AI teams), and becomes increasingly actionable as we learn more about the value of intuition and reflection.

## 1 Introduction

There was a time when it seemed like the most mentioned feature of language models was their size. A vestige of this size-obsessed era of generative transformer model development was the first ‘L’ in the acronym ‘LLM’. Consumer-friendly chatbots were *large*! But before chatbots were marketed for such broad use [61], terms like ‘language model’ and ‘GPT’ may have been preferred (see **Fig. 1**).

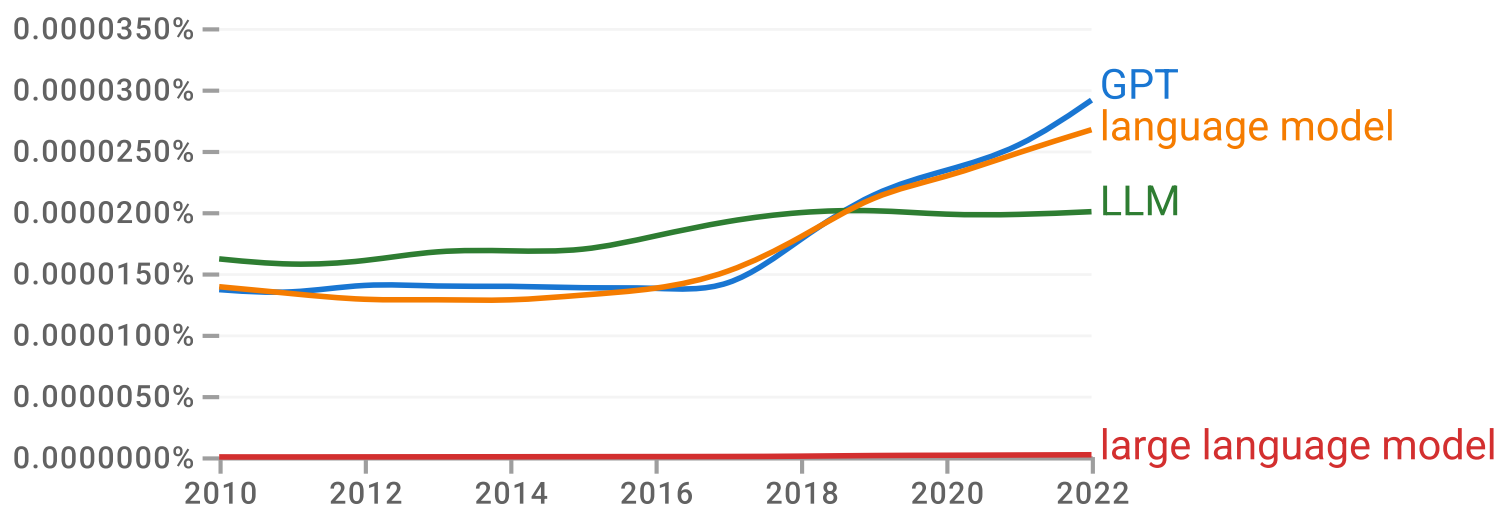

**Fig. 1.** Relative frequency of ‘large language model’, ‘LLM’, ‘language model’, and ‘GPT’ from 2010 to 2022 via Google nGram.

Accepted for publication in the proceedings of the 4th International Conference on Human and Artificial Rationalities, which are to appear in Lecture Notes in Computer Science.

Preprint available via ArXiv: https://doi.org/10.48550/arXiv.2505.22987

This paper will argue that — among other things — we can become too obsessed with factors such as size, chains of thought, and other features of reasoning models. While these features may be related to the reflective thinking that many want from models, they are merely means to our ultimate ends, which are often in tension. So what we really need from intelligent systems are optimal tradeoffs between these competing goals. This insight implies a kind of pragmatism about reasoning resources (in humans, machines, or human-machine collaborations). Completing the details of this position requires a review of some ideas from history and more recent scientific discoveries.

## 2 It's Not Their Size That Matters; It's How You Use Them

Language models do begin to outperform humans when they reach a certain size [30], indicating that there is *some* benefit to larger language models. But what is it about large language models that help them outperform humans?

In 2023, Eisape and colleagues reported that as models increased in size, they were more likely to "behave" like a system called mReasoner [44, 45], especially its propensity for System 2 or reflective thinking [30] (**Fig. 2**). Such results suggest that this capacity to reason reflectively probably matters at least as much as model size [12].

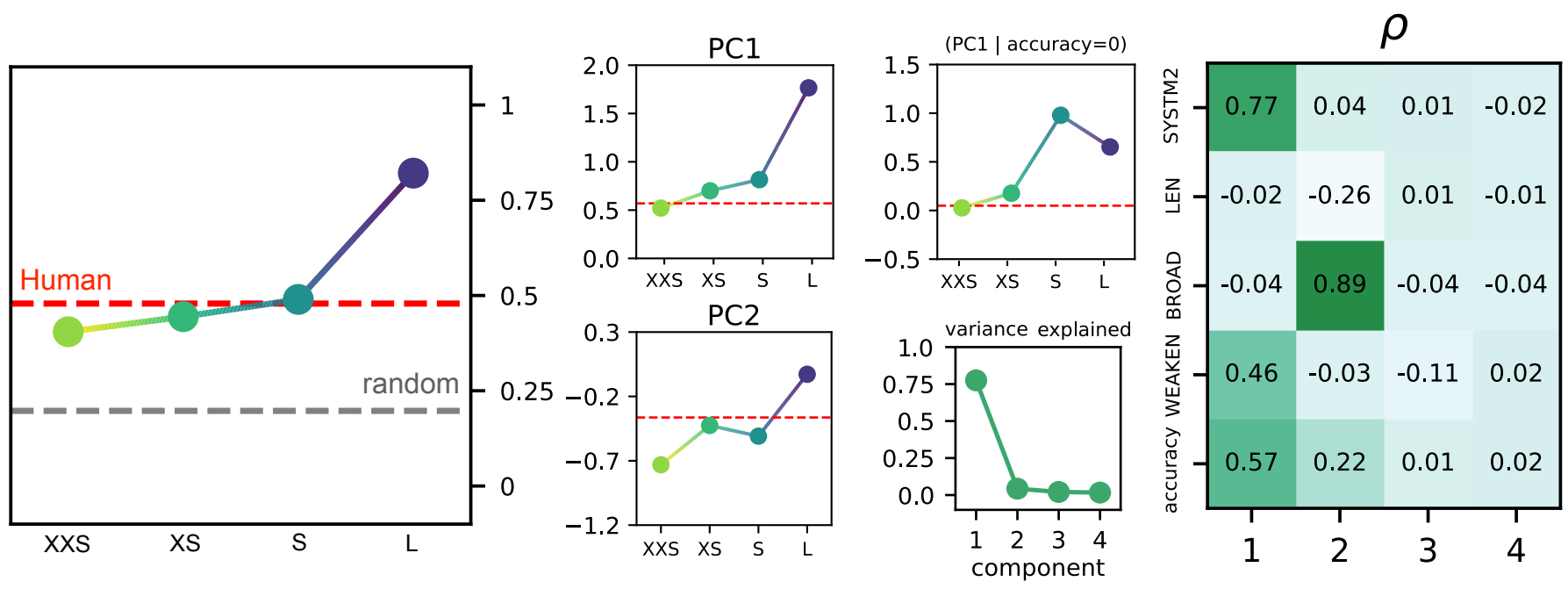

**Fig. 2.** Performance of language models on logic tasks by model size (left), the two principal components — PC1 and PC2 — that captured most of the variance in mReasoner's responses to the logic tasks (center), and correlations between these components of mReasoner's responses and its propensity to engage in System 2 or reflective thinking (right) from Eisape et al.

### 2.1 Reflection Tests

Cognitive scientists have developed behavioral tests of reflective reasoning [7]. The most famous reflection test question is the bat-and-ball problem [59]. "A bat and a ball cost $1.10 in total. The bat costs $1 more than the ball. How much does the ball cost?" [42]. Such reflection test questions are designed to lure people into accepting an alluring-yet-incorrect response (10 cents). So we call these tests of "System 2" or reflective thinking because they lure us into accepting a particular response that — with just a moment's reflection — we can realize is incorrect [72]. Moreover, some reflection tests seem to measure the two components that feature prominently in philosophers' descriptions of reflective thinking [18]: *deliberate* inhibition of one's initial intuition and conscious *awareness* of additional reasoning [14].

**Table 1.** True positive, false positive, true negative, and false negative rates for the 10-item verbal reflection test [70] distilled from humans' real-time verbalizations by Byrd et al. Example verbalizations are responses to the following item: "If you were running a race, and you passed the person in 2nd place, what place would you be in now?" [70]

| Category | Example Verbalization | Study 1 | Study 2 |
|---|---|---|---|
| Correct-and-reflective | "1st obviously. No actually . . . 2nd." | 80.2% | 68.5% |
| Correct-but-unreflective | "2nd" | 19.8% | 31.5% |
| Lured-and-unreflective | "1st" | 71.5% | 75.8% |
| Lured-yet-reflective | "I want to say 1st but, umm, yeah, 1st." | 28.5% | 24.2% |

## 2.2 Superficial Signs of Reflection

So-called "reasoning" models [5] appear to engage in multi-step reflection and often outperform humans on reflection tests [38]. These two facts may lead you to conclude that what distinguishes reasoning models from language models is a capacity for reflective thinking. However, much like humans often pass reflection tests without exhibiting any signs of reflective thinking [18, 75] (see "correct-but-unreflective" in **Table 1**), Hagendorf and colleagues found that language models continued to outperform humans on reflection tests even when they could not exhibit signs of reflective thinking, such as chain-of-thought reasoning [38] (**Fig. 3**).

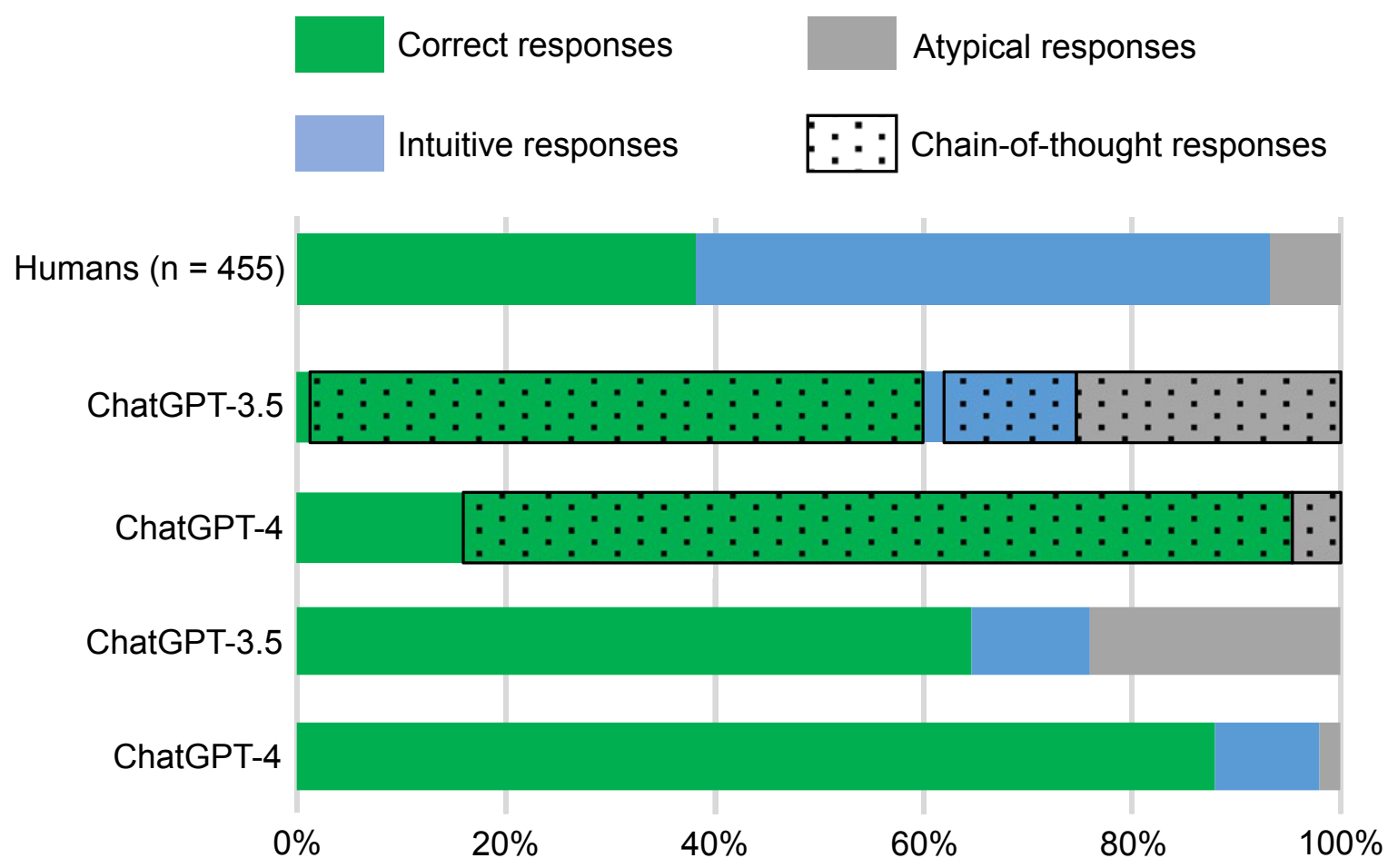

**Fig. 3.** Reflection test answers from humans and from language models, which were either free to use chain-of-thought (CoT) reasoning or else prevented from it (Hagendorf et al.).

Thus, seemingly reflective chains of thought can be merely performative [84, 57] or inefficient [54, 68, cf. 62]. In some cases, chains of thought (such as "wait, let me think about it") may even cause "bias" [81]. So, just as the size of language models can be overrated, so can *indirect* evidence of reflective thinking.

### 2.3 Romanticism About Reflection

Philosophers have long thought that reflective reasoning is crucial for good reasoning [9]. More recently, we have learned that reflective thinking predicts or causes changes in ethical and other philosophical judgments [15, 17, 19, 27], even among philosophers [11]. We are also learning that thinking about philosophy could make people think more reflectively [13]. But philosophers are not alone in their "preoccupation with reflection" [Chapter 2 of 29]. Nowadays, computer scientists are also arguing for the importance of reflection in reasoning models [46, 85].

Although there are benefits to reflectively stepping back from our initial impulses to become aware of more reasons and alternatives [14], plenty of good evidence suggests that reflective reasoning can also delay or even hinder decisions [10]. So, even if some of what we admire about human and artificial intelligence is a capacity for reflective reasoning, reflection can fail to fulfill our goals.

### 2.4 Idealized Rationality

We can also become so obsessed with rationality that we fail to recognize other important factors. Consider certain economic notions of rationality. Since at least John Stuart Mill, economists have been accused of overly idealized theories of human rationality, such as perfectly self-interested agents [60]. Of course, the history of the *homo economicus* ideal is more complicated [2], and even Mill admitted that no "political economist was ever so absurd as to suppose that mankind are really thus constituted" [60:322]. Nonetheless, a contingency of economic models seem to assume that human decisions seek to maximize or else continually increase utility of one sort or another.

By the 1950s, however, cognitive scientists like Herb Simon had argued that models should "take explicit account of the 'internal' as well as the 'external' constraints that define the problem of rationality for the organism" [69:101]. These arguments eventually led to models of "bounded rationality" [36], which can better capture how our decisions often fail to maximize or even increase utility [80, 82].

So even if we want reasoning models to be large, reflective, or rational, we must admit how various constraints can prevent them from optimizing these outcomes [22, 33]. There may also be constraints and circumstances that prevent us from wanting these outcomes from intelligent systems [56, 77]. Fortunately, some computer scientists have already realized the diminishing returns on further investment in the size of reasoning models, compared to the returns on more efficient architectures or inference techniques [1, 67].

### 2.5 Dual Systems Approaches To Intelligence

There is growing evidence in favor of a dual model approach, *a la* dual process or dual systems theory [32]. For example, Yan and colleagues found that pairing one small language model with another small model (that serves as a “reflective system”) allowed small hybrid systems to compete with larger models that had more than ten times as many parameters [86] (**Table 2**). These hybrid systems instantiate both aspects of reflection: rather than automatically accept output from the default system, a reflective system reasons about it further — the machine version of dual system psychology.

**Table 2.** Performance comparisons of standard, chain-of-thought, and tree-of-thought systems against smaller dual systems from Yan and colleagues.

| | | EB | | GSM8K | |
|---|---|---|---|---|---|
| **Model** | **#Params.** | **Accuracy (%)** | **Δ (%)** | **Accuracy (%)** | **Δ (%)** |
| **Comparative Systems** | | | | | |
| GPT-3.5 (code-davinci-002) | 175B | 80.76 | - | 16.17 | - |
| + Standard prompt | 175B | 84.23 | +3.47 | 17.03 | +0.86 |
| + Chain-of-thought prompt | 175B | 92.45 | +11.69 | 60.27 | +44.10 |
| + Tree-of-thought prompt | 175B | 93.31 | +12.55 | **61.39** | +45.22 |
| **CogTree** | | | | | |
| GPT2-XL (Intuitive System only) | 1.5B | 82.37 | - | 23.53 | - |
| + GPT2-XL (as Reflective System) | 1.5B | 92.63 | +10.26 | 35.84 | +12.31 |
| + LLaMA (as Reflective System) | 7B | 93.16 | +10.79 | 34.68 | +11.15 |
| LLaMA (Intuitive System only) | 7B | 86.14 | - | 43.52 | - |
| + GPT2-XL (as Reflective System) | 1.5B | 93.25 | + 7.11 | 47.80 | +4.28 |
| + LLaMA (as Reflective System) | 7B | **94.25** | +8.11 | 61.28 | +17.76 |

Others have also realized how pairing a small model with a larger model can offer “a promising trade-off between performance and computational efficiency” [87]. For example, Shang et al. found that multi-model systems can achieve “state-of-the-art reasoning accuracy and solution diversity” at a fraction of the token cost [66]. These results suggest that dual- or multi- model architectures may be key to yielding the reflective reasoning and rationality that we expect from intelligence systems, but without overlooking other goals or constraints [65].

Of course, deploying multiple language models does not guarantee higher performance at lower cost. Like model parameters, more models are not always better. So I am not proposing that we replace obsessive pursuit of model size or other factors with obsessive pursuit of additional models or maximally reflective inference. Rather, I am proposing a form of pragmatism [51].

## 3 Strategic Reflectivism: Pragmatism About Reflection

The core idea of this paper is that one key to intelligence is pragmatic (rather than perpetual) deployment of reflective reasoning— a view I have been calling *Strategic Reflectivism* [Section 4.2.3 in 10]. By tactically deploying reflective inference only when doing so is probably worth its costs, strategic reflectivism synthesizes pragmatism [73] with strategic reliabilism [3]. Strategic reflectivism is less optimistic about reflection than "reflectivism" (which holds that reflection is crucial for good reasoning [34]) yet more optimistic about reflection than anti-reflectivism (which argues that the benefits of reflection are illusory [29]). Strategic reflectivism has many implications. This paper will focus on the systems we study in cognitive science and computer science.

### 3.1 From Strategic Computation To Strategic Reflection

Consider an analogy with heterogeneous computing, which integrates multiple processing core types. For example, chips may contain both performance cores and efficiency cores, both general processing units and specialty processing units (such as graphics processing units or neural processing units). Such heterogenous system architectures can "outperform the best …homogeneous architecture" multiprocessors, often in less time, and with fewer resources [92].

Strictly speaking, strategic reflectivism is not committed to the claim that properly reflective inference requires multiple modules (or models, systems, cores, units, etc.). Reflection need only involve (1) deliberate suspension of the initial response and (2) awareness of additional reasoning about that response [14]. So just as one human can "back up" or "step back" from their initial response [76] and consciously answer a question [39] such as, "Shall I believe [that]?" [48], one machine can (in principle) iteratively prompt itself with questions about whether to endorse its own output [46]. Nonetheless, there are a few lessons we can draw about intelligent systems from the efficiency and performance of multi-core or multi-unit computing systems.

**Efficiency.** Like heterogenous computing systems, intelligent systems that can switch between default and more reflective inference can better optimize performance-efficiency tradeoffs than standard inference or always-reflective inference systems. For example, Sui and colleagues used a "Meta-reasoner" to evaluate a summary of each thought (in a chain) *before* executing any strategy; and their meta-reasoner method achieved better performance-cost tradeoffs than many other reflection protocols [74] (**Fig. 4**).

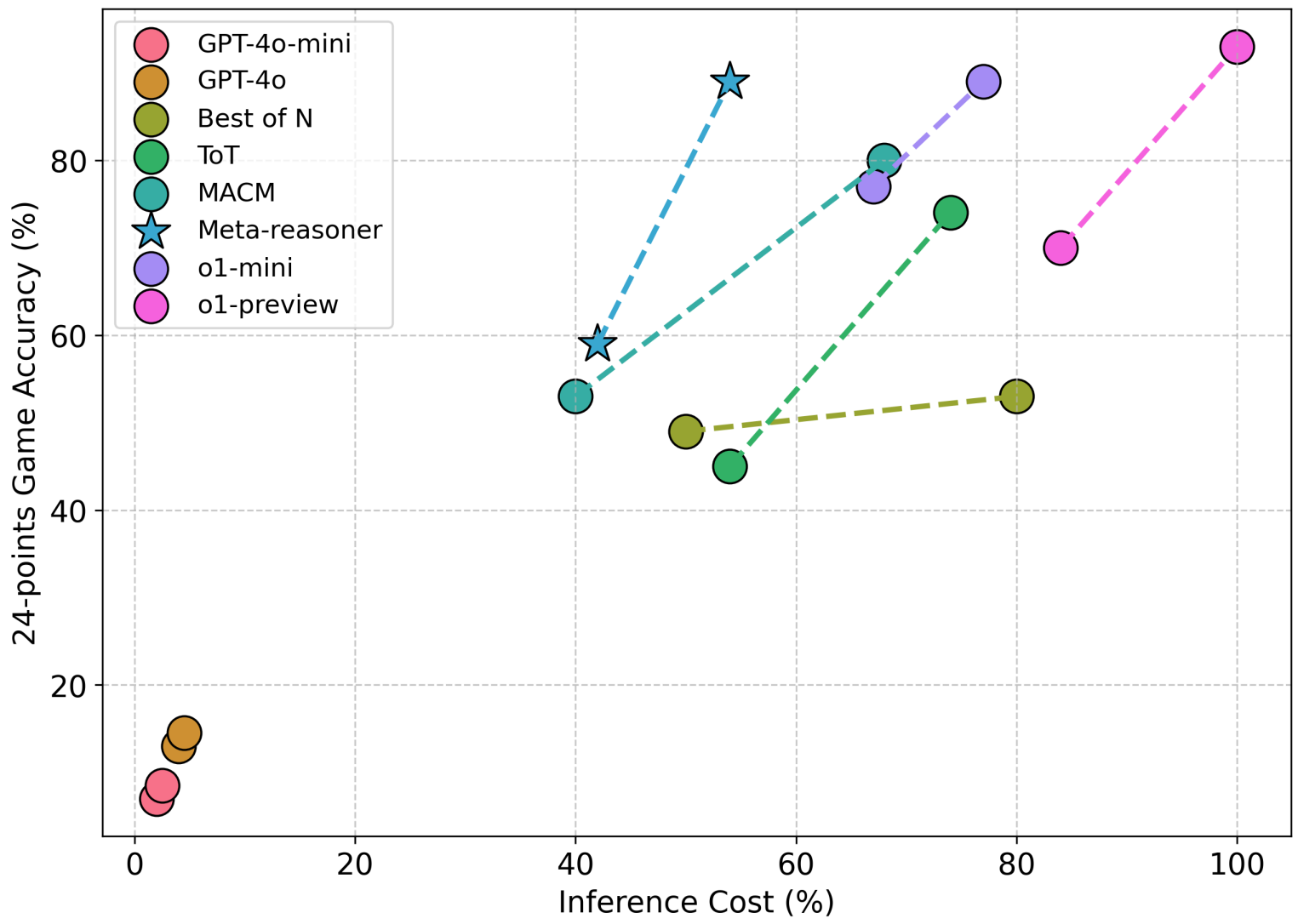

**Fig. 4.** Tradeoffs between performance and cost in various reasoning models, with Sui and colleagues' Meta-reasoner achieving the optimal trades.

**Pragmatism.** Heterogenous computing requires some sort of meta-process to determine — among other things — when to recruit additional cores or core types. Strategically reflective systems also need a method to efficiently allocate the resources necessary for reflective inference about default, intuitive responses [23]. Ideally this meta-process would function automatically and adaptively [21, 53, 53]. One approach might be an "introspective" Monte Carlo tree search that identifies and starts with the most promising nodes during inference [55]. This kind of meta-process may learn to let intuitive inference handle "common sense" or "general world knowledge" and thereby save resources for more tricky, multi-step tasks that are more likely to require reflective inference [89].

## 3.2 When To Start And Halt Reflection

As with computing, there are multiple ways to determine how to allocate resources in intelligent systems. Bounded Reflectivism synthesizes evidence from cognitive psychology to predict — among other things — when humans will (or should) reason reflectively [Section 3.3 in 10] (**Fig. 5**).

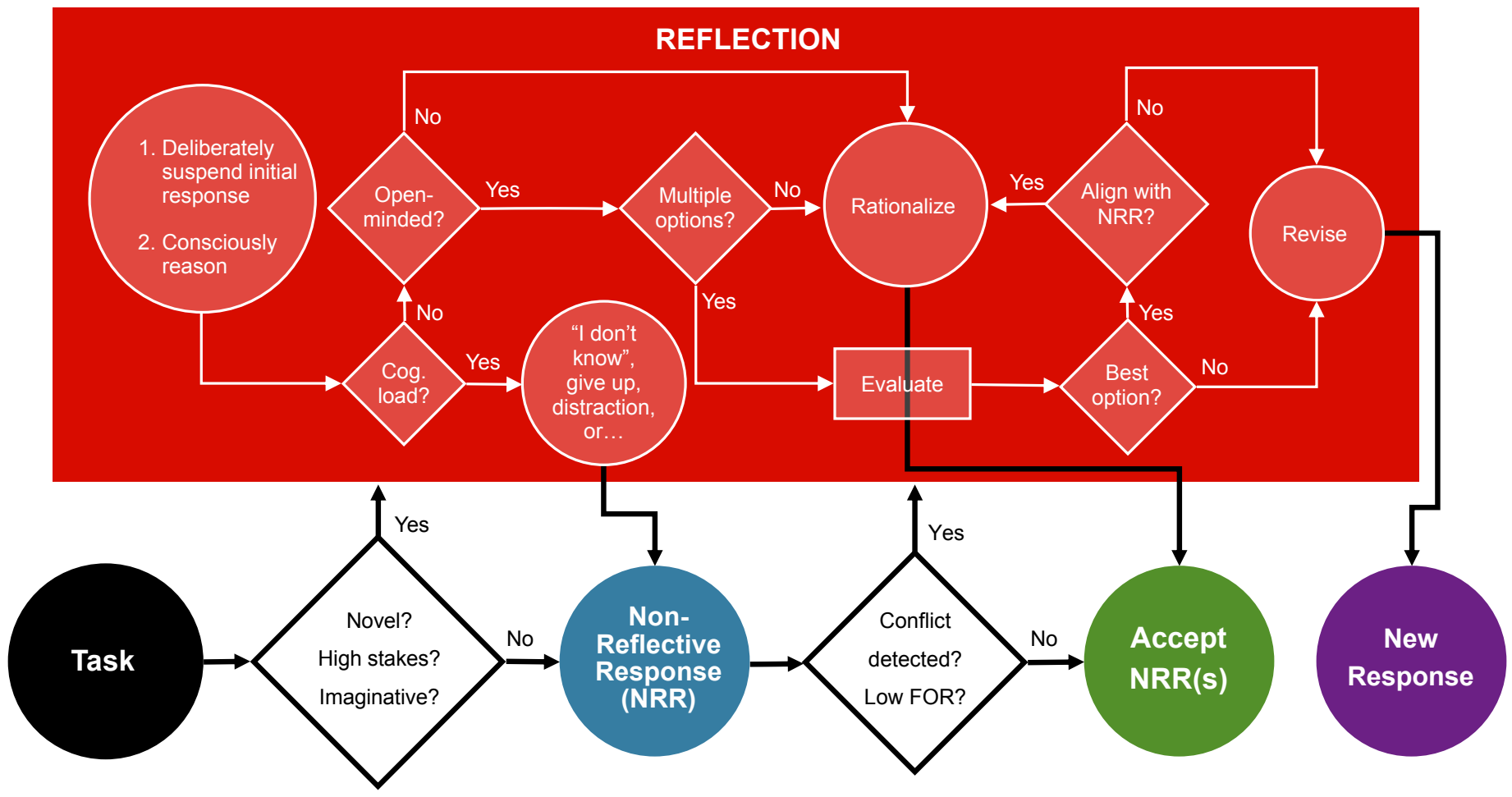

**Fig. 5.** The Bounded Reflectivism algorithm [10], according to which reflective thinking can be triggered by novel, high stakes, or imaginative tasks as well as responses that reveal conflict or yield low confidence, which is also known as low "feeling of rightness" (FOR) [78]. Variables such as open-mindedness and cognitive load can constrain the comprehensiveness and impartiality of reflective reasoning, *a la* bounded rationality.

Tasks that seem novel, high-stakes, or imaginative may nudge people to reflectively step back from our first impulse and consciously consider reasons or alternatives. We may also want to reflect when our first (intuitive or unreflective) response conflicts with our other beliefs or if we merely lack confidence in this initial response. Of course, reflection can be interrupted or even prevented by factors such as cognitive load or distraction. These insights from reflective thinking in humans may provide insight about other strategically reflective intelligent systems.

For example, when standard inference detects that a prompt or task requires knowledge that is unlikely to be in the system's training dataset, it would seem prudent to recruit another model or inference step to reflect on the default model's output. Alas, such out-of-distribution data is often undetected. So systems may also make decisions about whether to recruit a reflective model based on the confidence assigned to the default model's output — lower confidence scores would indicate opportunities for more reflective inference (depending on the stakes, of course). In principle, conflict detection could be handled by the default model (as in self-consistent chain-of-thought reasoning or in parallel processing [79]); however, given the aforementioned evidence about the superficiality (section 2.2) or inferiority (section 3.1) of some chain-of-thought reasoning, another reflective model or meta-process may be needed for optimal conflict detection [28].

Halting thresholds for reflective inference may also scale with stakes, familiarity, resource load, and other factors. For example, *a priori* budgets for inference cost, time, or other variables could prevent further reflection when reflective inference exceeds allotted resources [24]. Of course, more automatic and adaptive stopping rules would be ideal [52, 88].

There may be many more factors (besides familiarity, stakes, etc.) that should determine when to "intervene" on lower-cost, default inference with higher-cost reflective inference [31]. And some intelligent systems may determine when to start and stop a reflective model based on goals besides just performance and cost-savings [3, 73]. The main point of strategic reflectivism is that reflective reasoning (and its resources) should be allocated according to the system's goals, optimally fulfilling competing goals wherever possible. In short, strategic reflectivism treats reflective inference as a means, not an end — a tool, not a goal.

## 4 Strategically Reflective *Collective* Intelligence Systems

There is no need to restrict strategic reflectivism to one human or one artificial intelligence system. Collective reasoning may also benefit from strategic reflection [83, 64]. Indeed, humans who reflect together often perform better on tests of reflective thinking than those who reflect alone [16], raising-questions about strategic human-AI reflection. Perhaps the most straightforward strategically reflective human-AI team would involve a human as one system and a machine as the other system. Determining which one handles reflective inference may depend on the task, the human's expertise, etc. So human-AI reflection may take many forms. Consider the implications of strategic reflectivism for a few domains of human-AI teaming.

### 4.1 Debiasing

Concerns about bias in humans and machines are voluminous. Before the onset of popular, consumer-friendly AI systems, there were ongoing debates about how to measure and ameliorate human biases [7, 8, 20]. Many of these biases have been detected in machines as well [25, 40, 43]. Consider what the pragmatism of strategic reflectivism implies about how to deal with human and AI biases.

**Wisdom in unreflective crowds?** More than a century ago, Francis Galton noted how to yield more accurate estimates by merely aggregating the judgments of crowds, even when members of the crowd are not reasoning reflectively [35]. There is now evidence that human-AI aggregates yield better medical diagnoses than entirely human or entirely AI aggregates [90]. You might be thinking that such judgment aggregation does not constitute reflective inference — and that is part of the point. Aggregation is often less tedious or costly than reflective inference and the wisdom of an unreflective crowd is often good enough for many purposes. Indeed, aggregate polling results, stock market indexes, and weather forecasts are often informative and corrective *before* expert analysts reflect on them. So the strategic reflectivist should acknowledge that when judgment aggregation is easy, more reflective inference may not be worth its opportunity costs.

**Biased compared to what?** There is no shortage of public concern about bias in decision-making, especially when there are proposals to automate otherwise human decisions — *a la* algorithm aversion [49, 58, 63]. Alas, these public outcries often ne-glect opportunity costs. It may be true that the new regime of decision makers is imperfect; we may even know precisely how a new decision system is biased. Nonetheless, mere existence of a bias is not enough to dismiss a system or its conclusions [4, 6]. Strategic reflectivism reminds us that for complaints about bias to be actionable, they must demonstrate that a more reflective system is both viable and better at optimizing the tradeoffs between our competing goals. Until such criteria are met, a complaint about bias in intelligent systems may be no more potent than common complaints about collective checks and balances, to which Churchill famously responded: "democracy is the worst form of Government except all those other forms that have been tried" [91].

### 4.2 Decision Aids

Decision aids are designed to help people prevent mistakes, overcome bias, or improve their decisions more generally. And there is already evidence that non-experts seem to have benefitted from tools like chatbots when it comes to fact-checking [47, 26] and understanding their medical information [50]. These examples do not allow us to conclude that AI assistance will improve all decisions. Indeed, human medical experts assisted by GPT-4 have performed worse, on average, than both the language model itself and their unassisted colleagues [37]. So, insofar as those results generalize, strategic reflectivism may recommend employing AI-assisted reflective thinking more in non-experts than in experts. If these results do not generalize, then the lesson may just be that some decision aids are more effective than others.

### 4.3 Data Annotation

Making sense of unstructured data can require enormous human resources. In medicine, inexpert human-AI teams have more accurately annotated "low uncertainty" images than inexpert humans [41], which implies an opportunity to better allocate busy and expensive experts to "high uncertainty" images that are not as well annotated by the inexpert human-AI teams.

We can debate about how to classify those two image annotation pipelines as "intuitive" or "reflective". After all, when non-experts receive feedback from an AI system (or vice versa), they may reflectively "decouple" from their initial annotation to consider an alternative annotation inspired by the feedback [71]. And, in principle, both the human-AI team and the expert could competently complete some annotation tasks without reflectively backing up from every response to consider alternatives.

So the implication here is not that inexpert human-AI teams or human experts exhibit only one of the two systems from dual process theory (intuitive or reflective, System 1 or System 2); both groups can reason either intuitively or reflectively. Rather, the implication of this case is that strategically reflective systems may decide whether to deploy their reflective reasoner (human or AI) based on factors such as the task domain, the default reasoners' expertise in that domain, or some combination thereof.

## 5 Conclusion

The first key claim in this paper was that reasoning *architecture* can matter at least as much to the success of intelligence systems as factors such as size [12], chains of thought, or other superficial indicators of reflection. Indeed, empirical studies of language models often find that properly dual-model systems can be more efficient than larger single-model systems. However, history reminded us that resource-intensive reasoning is not always worth it and maximizing one form of utility is unrealistic. Insofar as we care about real-world decisions and actions, we want some kind of pragmatism.

So *strategic reflectivism* proposed pragmatic (rather than permanent) deployment of reflective inference. Like the performance and efficiency cores of heterogenous computing systems, systems capable of switching between more and less reflective inference can outperform standard inference systems and perpetually reflective systems, often with less cost and time.

Importantly, strategic reflectivism is agnostic about reasoning content and its reasoner. Reflection can be unimodal or multimodal — and it can occur in humans, machines, or collections thereof. So the view's insight and normative implications are actionable for nearly any field that develops, studies, or uses intelligent systems: business, computer science, decision science, education, epistemology, healthcare, intelligence analysis, policymaking, etc.

Finally, strategic reflectivism remains actionable even if the empirical results cited herein fail to replicate or generalize. Indeed, the more we learn about the contextual value of intuitive and reflective reasoning, the more strategic reflectivism can advise its stakeholders.

**Acknowledgments.** Thanks to Herb Simon and Steve Stich for advancing American pragmatism and to reviewers (human and machine) for comments. An earlier shorter version of this paper was presented at the 2025 ACM Workshop on Human-AI Interaction for Augmented Reasoning (AIREASONING-2025-01). 

**Disclosure of Interests.** I am not aware of competing interests that are relevant to the content of this article. In case conflicts are unwitting, my research has been funded by the U.S. Office of the Director of National Intelligence, the U.S. Department of Energy, the John Templeton Foundation, and CloudResearch.